\documentclass[unnumsec,webpdf,modern,large,numbib]{oup-authoring-template}

\usepackage{pifont}
\usepackage{microtype}
\usepackage{inconsolata}
\usepackage{subfig}
\usepackage{bold-extra}
\usepackage{bm}
\usepackage[hang, flushmargin]{footmisc} 
\usepackage{tabularray} 
\NewColumnType{M}{Q[c,m]}
\NewColumnType{W}[1]{Q[l,m,wd=#1]}
\NewColumnType{L}{Q[l,m]}
\NewColumnType{N}[1]{Q[c,m,wd=#1]}
\newcommand{\rtb}[1]{\rotatebox[origin=c]{90}{#1}}
\usepackage[most]{tcolorbox}
\definecolor{cpurple}{RGB}{147, 145, 255}
\definecolor{cblue}{RGB}{109, 177, 255}
\definecolor{cframe}{RGB}{0, 61, 135}
\definecolor{cfill}{RGB}{237, 245, 255}
\definecolor{cpink}{RGB}{217, 22, 168}
\definecolor{cgreen}{RGB}{0, 194, 168}
\definecolor{mname}{RGB}{140, 148, 158}
\definecolor{num}{RGB}{104, 104, 104}
\definecolor{optional}{RGB}{218, 223, 228}
\newtcbox{\dbox}[1][]{ on line, enhanced, frame hidden, interior hidden, sharp corners, size=fbox, borderline={0.4pt}{0pt}{dashed}, #1 }
\newtcbox{\bbox}[1][]{ on line, enhanced, frame hidden, size=small, arc=2mm, #1 }
\newtcbox{\fcbox}[1][]{ on line, enhanced, rounded corners, size=fbox, #1 }
\newtcbox{\cbox}[1][]{ on line, enhanced, frame hidden, sharp corners, size=fbox, #1 }

\newcommand{\para}[1]{\vspace{0.3em}\noindent\textbf{#1}$\:$ }

\usepackage{tikz}
\usetikzlibrary{shapes.geometric, arrows.meta, positioning, calc}

\graphicspath{{Fig/}}

\theoremstyle{thmstyleone}%
\theoremstyle{thmstyletwo}%
\theoremstyle{thmstylethree}%

\begin{document}

\journaltitle{Journal Title Here}
\DOI{DOI HERE}
\copyrightyear{2022}
\pubyear{2019}
\access{Advance Access Publication Date: Day Month Year}
\appnotes{Paper}

\firstpage{1}


\title[TRUST Dialogue System]{TRUST: An LLM-Based Dialogue System for Trauma Understanding and Structured Assessments}

\author[1,$\ast$]{Sichang Tu}
\author[2]{Abigail Powers}
\author[3]{Stephen Doogan}
\author[1]{Jinho D. Choi}

\authormark{Tu et al.}

\address[1]{\orgdiv{Department of Computer Science}, \orgname{Emory University}, \orgaddress{\street{Atlanta}, \postcode{GA 30322}, \country{United States}}}
\address[2]{\orgdiv{Department of Psychiatry and Behavioral Sciences}, \orgname{Emory University}, \orgaddress{\street{Atlanta}, \postcode{GA 30322}, \country{United States}}}
\address[3]{\orgname{DooGood Foundation}, \orgaddress{\street{Miami}, \postcode{FL 33179}, \country{United States}}}

\corresp[$\ast$]{Corresponding author. \href{email:sichang.tu@emory.edu}{sichang.tu@emory.edu}}

\received{Date}{0}{Year}
\revised{Date}{0}{Year}
\accepted{Date}{0}{Year}

\abstract{
    \textbf{Objectives:} 
    While Large Language Models (LLMs) have been widely
    used to assist clinicians and support patients, no existing work has explored
    dialogue systems for standard diagnostic interviews and assessments. This study aims
    to bridge the gap in mental healthcare accessibility by developing an LLM-powered
    dialogue system that replicates clinician behavior. \\
    \textbf{Materials and Methods:} We introduce TRUST, a framework of cooperative LLM modules
    capable of conducting formal diagnostic interviews and assessments for Post-Traumatic Stress Disorder (PTSD).
    To guide the generation of appropriate clinical responses,
    we propose a Dialogue Acts schema specifically designed for clinical interviews.
    Additionally, we develop a patient simulation approach based on real-life interview transcripts
    to replace time-consuming and costly manual testing by clinicians.\\
    \textbf{Results:} A comprehensive set of evaluation metrics is designed to assess the dialogue system
    from both the agent and patient simulation perspectives.
    Expert evaluations by conversation and clinical specialists show that
    TRUST performs comparably to real-life clinical interviews.\\
    \textbf{Discussion:} Our system performs at the level of average clinicians,
    with room for future enhancements in communication styles and response appropriateness.\\
    \textbf{Conclusions:} Our TRUST framework shows its potential to facilitate mental healthcare availability.
}



\maketitle

\section{Introduction}
\label{sec:introduction}

Over 28 million adults with mental illness in the United States do not receive any treatment,
with more than half reporting unsuccessful attempts to access care.\citep{reinert2023state}
Many of them are unable to take the first step toward the treatment - getting formal diagnosis,
due to critical shortage of mental health providers and prohibitive medical cost.
Structured diagnostic interviews based on the
Diagnostic and Statistical Manual of Mental Disorders (DSM-5)\citep{american2013diagnostic}
are considered the gold standard for accurately diagnosing mental health conditions.
However, these assessments require substantial clinical expertise and time investment,
creating further bottlenecks in mental healthcare delivery systems.
Post-Traumatic Stress Disorder (PTSD) presents a particularly concerning case among common mental health conditions,
as it often goes undetected or overlooked in civilian populations.
For instance, trauma exposure exceeds 85\% in safety-net hospitals,\citep{Gillespie2009, Gluck2021}
yet comprehensive trauma assessments remain limited across many healthcare settings,
which widens the care gap for individuals affected by PTSD.

Recent advances in Large Language Models (LLMs) have demonstrated their capabilities
in conducting contextually appropriate conversations,\citep{Xi2023}
leading to active exploration of their applications in mental healthcare dialogue systems
for condition detection,\citep{Chen2023} intervention,\citep{Jo2023} and counselling.\citep{liu2023chatcounselor}
However, few studies have focused on diagnostic dialogue systems.
Closest to our work, \citet{Chen2023a} explored the feasibility of using ChatGPT
to simulate both doctors and patients through prompt engineering guided by psychiatrists.
Existing research primarily addresses more prevalent mental health conditions,
such as depression and anxiety disorders, within short conversations.
Notably, there remains a significant gap in investigating formal diagnostic dialogue systems capable of conducting time-intensive structured clinical interviews while simultaneously performing diagnostic assessments.


\begin{figure*}[ht!]
    \centering
    \includegraphics[width=\textwidth]{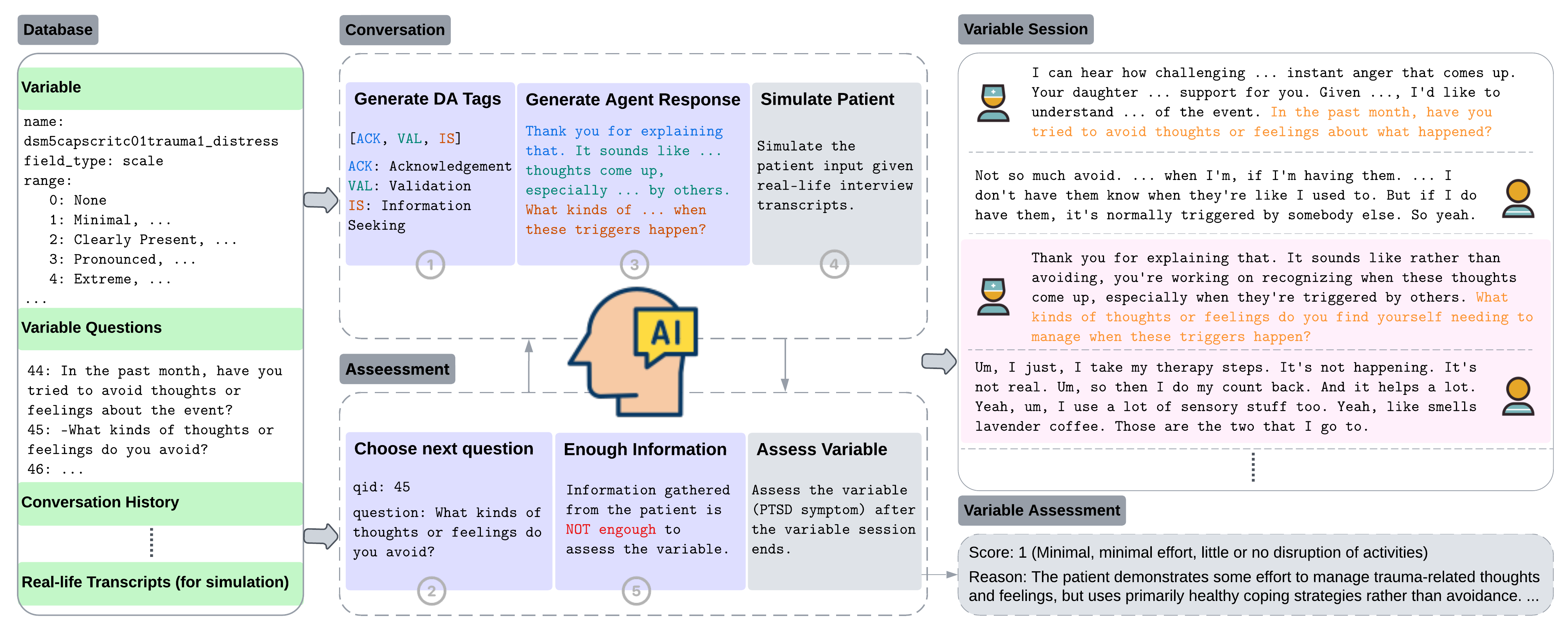}
    \caption{Overview of the TRUST system in an example variable session.
        \fcbox[colframe=mname,colback=mname]{Database} stores metadata for each variable (e.g., symptoms to be assessed).
        \fcbox[colframe=mname,colback=mname]{Conversation} and \fcbox[colframe=mname,colback=mname]{Assessment}
        are LLM-powered cooperative AI modules that control dialogue flow, generate clinician responses, and perform assessments.
        These modules use variable metadata and follow the numbered steps \textcolor{num}{\ding{172}-\ding{176}} to generate the conversation,
        with grey steps indicating optional actions.
        \fcbox[colframe=optional,colback=optional]{Simulate Patient} is used only during system evaluation,
        and \fcbox[colframe=optional,colback=optional]{Assess Variable} is triggered once sufficient information has been collected.
        \fcbox[colframe=mname,colback=mname]{Variable Session} illustrates a generated conversation for the example variable,
        which assesses the intensity of the patient's avoidance of trauma-related thoughts or feelings in the past month.
        \fcbox[colframe=mname,colback=mname]{Variable Assessment} presents the corresponding assessment result predicted by the system.}
    \label{fig:overview}
\end{figure*}

To bridge this research gap, we present TRUST, a LLM-powered dialogue system for
\textbf{TR}auma \textbf{U}ndersatnding and \textbf{ST}ructured Assessments,
designed to conduct comprehensive formal diagnostic interviews and assessments
for PTSD in accordance with the Clinician-Administered PTSD Scale for DSM-5.\citep{weathers-2018}
Figure~\ref{fig:overview} gives the overview of TRUST in an example variable session.
Our system has been evaluated by both conversation expert and clinical specialist
familiar with PTSD interviews,
with results indicating equivalent performance to real-life clinical interviews.
Our primary contributions include:\footnote{Our system, prompt, and examples of generated transcripts are publicly available through our open-source project at \url{https://github.com/SichangTu/TRUST.git}.}

\begin{itemize}
    \setlength\itemsep{0em}
    \item A Dialogue Acts (DA) schema for clinician utterances in diagnostic interviews
          that serves as an intermediate layer to improve system controllability
          and could generalize to various mental health conditions.
    \item An innovative dialogue system that acts as an expert clinician to conduct hour-long formal diagnostic interviews and assessments for PTSD.
    \item A novel approach to simulate patient responses for robust system evaluation using real-life clinician-patient interview transcripts as the simulation reference.
\end{itemize}

To our knowledge, TRUST is the first diagnostic dialogue system tailored to
conduct formal diagnostic interviews mirroring real-world clinical practice.
It offers potential benefits for both mental healthcare recipients
by reducing cost barriers and stigma associated with formal diagnosis,
and for healthcare providers by decreasing time investments in interviews and screenings,
thereby enabling them to serve more patients effectively.
This advancement is particularly valuable for conditions like PTSD,
where the combined challenges of mental health provider shortages and
limited specialized trauma training create significant barriers to care.
Importantly, we develop TRUST following general design principles that enhance its adaptability.
It allows the system to be modified for other mental health conditions when structured interview protocols are available,
extending its potential impact beyond PTSD to address broader mental healthcare needs.

\section{Methods}
\label{sec:methods}

\subsection{Data}
\label{ssec:data}

We utilize the PTSD interview dataset developed by \citet{tu-etal-2024-automating},
which contains transcribed and processed recordings of real-life diagnostic interviews between clinicians and patients.
The dataset encompasses four interview sections.
In this paper, we focus on the Clinician-Administered PTSD Scale for DSM-5 (\texttt{CAPS}) section.
\texttt{CAPS} represents the standard diagnostic criteria for PTSD as outlined in the DSM-5 guidelines,
and accounts for 92 out of the total 147 variables in the complete dataset.


\begin{table}[htbp!]
    \centering 
    \caption{Statistics of the \texttt{CAPS} section in the PTSD interview dataset and subsets.
        Original refers to the full dataset introduced by \citet{tu-etal-2024-automating}.
        Sampled represents the subset of transcripts selected for evaluating TRUST.
        Annotated indicates the portion of the data manually labeled with DA tags.}
    \label{tab:dataset}
    \begin{tblr}{
        colspec = {MMMMM},
        hline{1, Z} = {0.5pt, solid},
                hline{2} = {0.3pt, solid},
            }
                  & \textbf{Count} & \textbf{Turns} & \textbf{Sents} & \textbf{Tokens} \\
        Original  & 335            & 39,030         & 227,263        & 2,430,606       \\
        Sampled   & 100            & 10,735         & 55,813         & 596,557         \\
        Annotated & 5              & 1,242          & 6,056          & 78,024          \\
    \end{tblr}
\end{table}

In the \texttt{CAPS} section, each variable corresponds to a specific PTSD diagnostic criterion that
clinicians must assess during the interview process.
The variables are structured according to the DSM-5 manual,
with each variable potentially associated with multiple predefined interview questions.
These questions, also sourced from the DSM-5 manual, guide clinicians in gathering necessary information
to assess each diagnostic criterion.
Of the 92 variables in the \texttt{CAPS} section, some require no direct questions while others necessitate
multiple inquiries, resulting in a total of 241 predefined interview questions.

For evaluation, we sample 100 \texttt{CAPS} section transcripts from the original dataset.
Table~\ref{tab:dataset} shows the comparative statistics between the original \texttt{CAPS} section data and our sampled subset.

\subsection{Dialogue Acts}
\label{ssec:dialogue_acts}

Diagnostic interviews present unique challenges compared to open-domain dialogue systems,
as they must maintain the balance between natural conversation flow and structured diagnostic assessment
that adheres to clinical protocols.
In real-life clinical interviews, clinicians face complex decision points:
they must appropriately acknowledge patient experiences, determine when to probe further with follow-up questions,
select contextually appropriate questions, and manage transitions between diagnostic topics.
These requirements make it particularly challenging for end-to-end dialogue systems in the healthcare domain,
as they offer limited control over intermediate decision-making processes and
may generate responses that are inappropriate or potentially risky in diagnostic contexts.


\para{DA Tags}
Dialogue acts represent the functional role of an utterance within a conversation.
While widely used in modeling both open-domain and task-oriented dialogues,\citep{Godfrey1992,Boyer2010}
research on dialogue act schemas in healthcare settings remains limited.\citep{Perez-Rosas2016,Guntakandla2018,Bifis2021,Xu2023,Farzana2020}
To address these challenges, we develop a Dialogue Acts schema tailored to clinical interviews.
This approach decomposes complex decision-making into manageable steps,
enabling better control over conversation flow and systematic tracking of the interview state.
Through analyzing clinician responses from real-life interview transcripts,
we conclude 8 tags:

\begin{itemize}
    \setlength\itemsep{0em}
    \item \texttt{GC} (Greeting/Closing) initiates or concludes the interview.
    \item \texttt{GI} (Guidance/Instructions) offers instructions, educational information, or procedural guidance.
    \item \texttt{ACK} (Acknowledgment) provides brief verbal confirmation of hearing or understanding patient input (e.g., \textit{I see}, \textit{Okay}).
    \item \texttt{EMP} (Empathy/Support) demonstrates understanding of patient's emotions or experiences (e.g., \textit{That must be difficult}).
    \item \texttt{VAL} (Validation) restates or summarizes patient's input to verify accurate understanding.
    \item \texttt{IS} (Information-Seeking) ask predefined assessment questions from the clinical protocol.
    \item \texttt{CQ} (Clarification Questions) requests additional details or explanation about patient's previous response; or invites patient to respond or elaborate.
    \item \texttt{CA} (Clarification Answers) responds to patient questions with clinical information, or procedural guidance.
\end{itemize}
These tags are not specific to PTSD and can potentially be applied to other clinical scenarios involving structured diagnostic interviews.

\para{DA Tag Annotation}
Two annotators with deep familiarity with the PTSD interview dataset
conduct annotation of 5 interview transcripts.
To ensure the generalizability of the DA tags,
we include all four interview sections from the original dataset in the annotation process.
Table~\ref{tab:dataset} provides the annotation statistics.
Our analysis yields an Inter-Annotator Agreement score of Cohen's kappa\citep{cohen1960coefficient} at 0.85,
indicating almost perfect agreement between annotators.


\para{DA Tag Prediction}
We employ Claude\footnote{We use \texttt{claude-3-5-sonnet-20241022}
    for all present experiments in this paper.
    We choose Claude as it is natively supported by the Amazon Web Services (AWS)
    that will host our diagnostic dialogue system.
    In principle, any Large Language Model could be utilized within our system.}
to generate the most appropriate 1-3 tags
for the next clinician response based on previous interview history.
Due to the inherent class imbalance in our dataset,
where tags such as \texttt{GC} and \texttt{CA} appear less frequently than dominant tags like \texttt{ACK} and \texttt{IS}, we utilized micro F1 score as our primary evaluation metric.
The model achieves an F1 score of 0.63 across the prediction task.
Through analysis of prediction results, we identified two significant challenges:
1) real-life conversations often feature casual and fragmented exchanges
where the model generates multiple dialogue act tags (e.g., \texttt{ACK}, \texttt{VAL}, and \texttt{IS})
when the gold standard indicates only \texttt{ACK}, with \texttt{VAL} and \texttt{IS} appearing in subsequent clinician responses.
2) the model demonstrates a tendency to overproduce clarifying questions (\texttt{CQ}),
sometimes in combination with information-seeking (IS) tags,
especially when patient responses are brief.

\begin{figure*}[ht!]
    \centering
    \includegraphics[width=\textwidth]{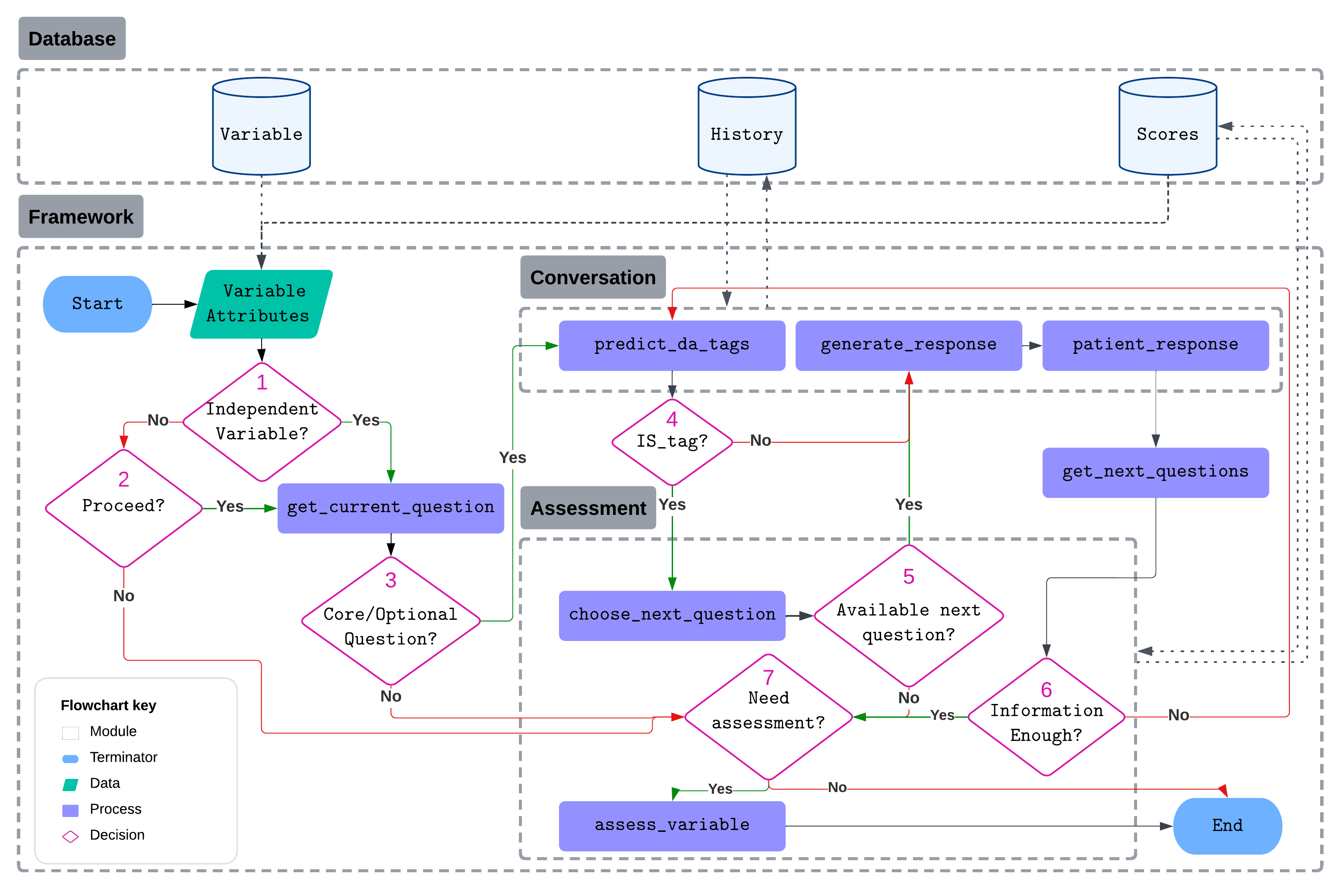}
    \caption{Dialogue flowchart design for each variable. This architecture integrates \dbox{Modules}
        for \fcbox[colframe=cframe,colback=cfill]{Database} interaction, conversation
        management and assessment logic to ensure adaptive and contextually appropriate
        interactions tailored for individuals. \fcbox[colframe=cpink, colback=white]{Decision node} represents key decision points, while \bbox[colback=cpurple]{Process node} indicate processing functions that manage dialogue generation and assessment tasks.}
    \label{fig:framework}
\end{figure*}

\subsection{Diagnostic Interview System}
\label{ssec:dialogue_system}

Due to the structured nature of diagnostic interviews, we design the dialogue system
with interconnected modules that cooperate to follow specific diagnostic
criteria while maintaining natural conversation flow.
Figure~\ref{fig:framework} presents the flowchart for the system design.
Our dialogue system contains two main modules: \textsc{Database} and \textsc{Framework}.
The \textsc{Framework} module contains two LLM-powered submodules,
\textsc{Conversation} and \textsc{Assessment}, which work in tandem to guide the conversational
directions and generate contextually appropriate responses.

\subsubsection{Database}
\label{sssec:database}

The \textsc{Database} module serves as the system's memory and is initialized
with three components: \texttt{Variable}, \texttt{History}, and \texttt{Score}.

\begin{itemize}
    \setlength\itemsep{0em}
    \item \textbf{\texttt{Variable}}
          The component stores comprehensive data about each variable in the \texttt{CAPS} section,
          including variable dependencies, associated interview questions, and metadata.
          Variables are categorized as either independent or dependent.
          Independent variables are essential elements that cannot be skipped during the interview,
          and their assessment does not rely on previous variable results.
          Dependent variables are assessed when previous patient responses indicate certain conditions
          that require further diagnosis, and their assessment may incorporate findings from prerequisite variables.
          Interview questions within each variable are organized in a hierarchical tree structure and
          are classified as either core or optional questions.
          Core questions are fundamental inquiries that must be asked during the interview.
          Optional questions serve as predefined follow-up inquiries that are posed when additional clarification
          or information is needed.
          The variable metadata encompasses variable type specifications, value ranges, assessment scales,
          special assessment conditions, and keywords used in replaceable prompt templates.
    \item \textbf{\texttt{History}} The component maintains a detailed record of each conversation turn,
          capturing temporal information (timestamps), contextual identifiers (variable ID, question index, and DA tags),
          and interaction content (system responses and patient inputs).
    \item \textbf{\texttt{Score}} The component serves as a repository for assessment outcomes for each variable,
    systematically storing variable identifiers, generated assessment reasoning and  scores.

\end{itemize}

Different from the \texttt{Variable} component, the \texttt{History} and \texttt{Score} components begin
as empty repositories that update dynamically during the interview if not resuming from an existing session.
These components actively exchange information with the \textsc{Framework} module, enabling the system to
maintain conversation coherence and make informed decisions about interview progression.

\subsubsection{Framework}
\label{sssec:framework}
The \textsc{Framework} module, as the core of our system,
comprises two submodules: \textsc{Conversation} and \textsc{Assessment},
which manage dialogue flow and diagnostic assessment.
It employs a sophisticated dialogue management approach
to handle each variable session.

Each variable session begins with the initialization phase, where the system retrieves variable metadata,
associated interview questions, and previous assessment results when applicable.
The system then examines the current variable through a series of decision points.
For independent variables, the system directly accesses the current node in the interview questions tree structure
\ding{172}\footnote{The number corresponds to decision points in Figure~\ref{fig:framework}.}.
For dependent variables, the system evaluates whether further investigation is warranted based on
assessment scores from prerequisite variables \ding{173}.

The interview process proceeds based on the availability of core or optional questions at the current node \ding{174}.
Some variables, while not associated with direct interview questions,
require assessment based on previous interview history and are routed directly to the \textsc{Assessment} module.
Variables that requires questioning are directed to the \textsc{Conversation} module with
the available interview questions from the current node and previous interview history.

The variable dialogue flow begins with the \texttt{predict\_da\_tags} function,
which generates appropriate DA tags.
These tags guide response generation in two ways \ding{175}: when tags exclude the \texttt{IS} (Information Seeking) tag,
the module generates contextual responses; when tags include the IS tag, indicating the need for predefined
interview questions, the \texttt{choose\_next\_question} function selects appropriate questions from the available options.
If next questions are available, the system generates the clinician response;
otherwise, it transitions to the variable assessment process \ding{176}.

Patient responses could either be manual patient input or
simulated patient response given the real-life interview transcripts.
Following patient responses, the \texttt{get\_next\_questions} function retrieves
remaining available questions for the current variable.
The \textsc{Assessment} module then evaluates whether the existing interview
history provides sufficient information for diagnostic assessment  \ding{177}.
This evaluation is crucial as patients may provide comprehensive responses
that address multiple inquiry points simultaneously, potentially eliminating the need for subsequent questions.

If the information is deemed insufficient, the system initiates another iteration of the dialogue loop,
beginning with dialogue tag prediction.
The process continues until either no further questions are available or sufficient information has been gathered.
At this point, the module determines whether variable assessment is necessary \ding{178}.
Variables may bypass assessment if they are section instructions or transitions,
or if they were previously assessed in a resumed session.
The \texttt{assess\_variable} integrates the automated pipeline for PTSD assessments developed by \citet{tu-etal-2024-automating}.

\subsection{Patient Simulation}
\label{ssec:simulation}




\begin{figure*}[htbp!]
    \centering
    \includegraphics[width=\textwidth]{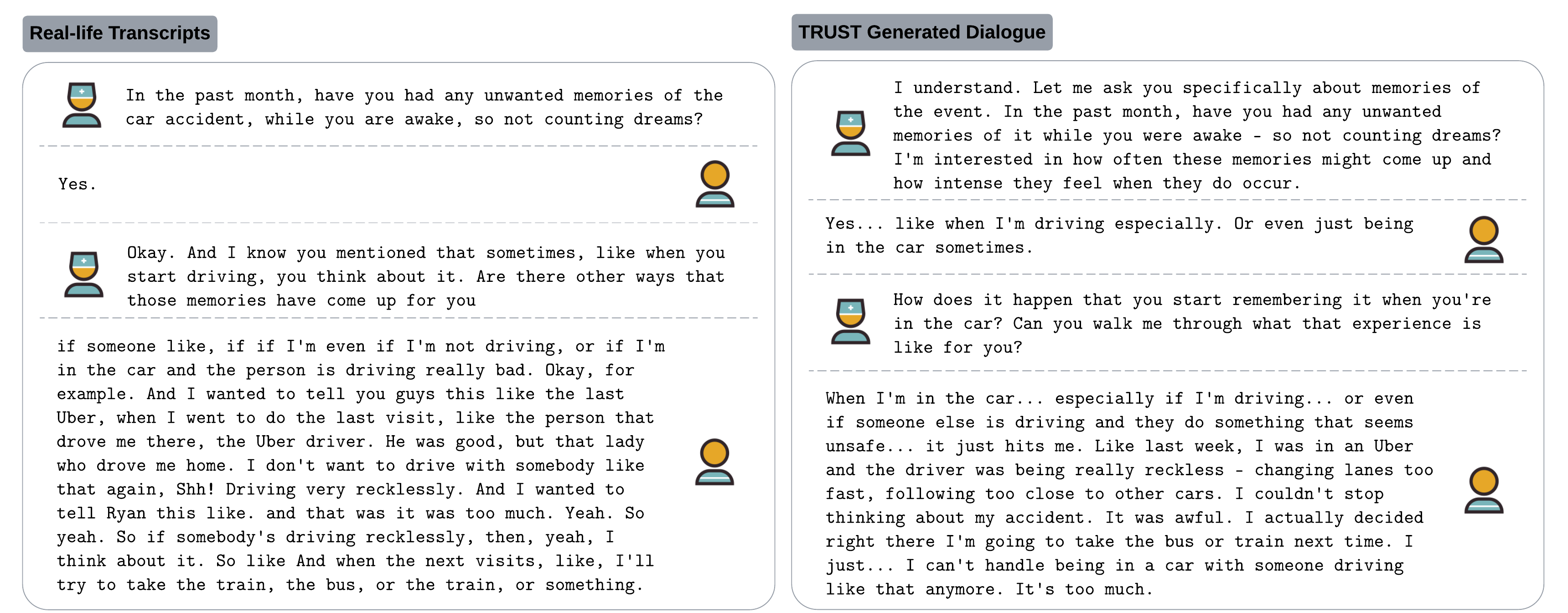}
    \caption{An example of patient simulation in TRUST.
        The left side shows segments from a real-life PTSD interview transcript,
        while the right side presents the generated dialogue produced by TRUST.
        Both the clinician and patient responses on the right side are generated by our system.}
    \label{fig:simulation}
\end{figure*}

Robust testing is critical for any clinical application before deployment with actual patients.
However, this poses significant logistical challenges.
Clinical evaluation by our research team is limited in scale,
and manual testing is both time-consuming and impractical.
Having clinicians repeatedly input responses and role-play different patient persona
with varied traumatic experiences if neither efficient nor representative of realistic clinical scenarios.

To address these constraints, we develop a novel patient simulation approach.
There is an increasing interest in leveraging LLMs to simulate various roles for professional training,
particularly in domains such as education \citep{Markel2023} and counseling \citep{Chen2023a}.
\citet{louie-etal-2024-roleplay} introduced Roleplay-doh, a pipeline that
incorporates domain expert feedback to simulate patients and train novice counselors in mental health settings.
Similarly, \citet{Wang2024} developed Patient-$\Psi$,
which employs cognitive modeling to simulate patients for cognitive behavioral therapy (CBT) training.



Our approach is distinguished by its foundation in real-life clinical data.
Rather than relying on theoretical models to simulate patient responses,
our system leverages actual interview transcripts as the source material for patient simulation.
This enables large-scale, clinically authentic system testing across diverse patient presentations
without requiring continuous clinician involvement.
Figure~\ref{fig:simulation} presents an example of the patient simulation approach.

The simulation operates on a variable-by-variable basis,
extracting relevant segments from the original transcripts to inform response generation.
When transcript data is available, the LLM is prompted to generate adaptive responses
that preserve core clinical information while maintaining a natural and realistic tone.
In cases where the transcript lacks relevant content for a specific interview question,
the model is instructed to respond with appropriate expressions of uncertainty
(e.g., \textit{I don't know} or \textit{I am not sure}) to minimize the risk of hallucinating clinical details.
To enhance response authenticity, we explicitly instruct the model to emulate
the patient's communication patterns observed in the original transcripts.
This includes replicating the granularity of information disclosure, maintaining consistent pacing
and hesitation patterns, and appropriately representing any topic avoidance or discussion difficulties
characteristic of PTSD patients.

\section{Results}
\label{sec:Results}

\subsection{Evaluation}
\label{ssec:evaluation}

To evaluate our system from diverse and objective perspectives,
we conduct human evaluation involving both conversation and domain experts.
The conversation expert possesses a linguistics background,
while the domain expert is a board-certified psychologist specializing in PTSD.
Both evaluators are familiar with the interview process and predefined questions
within the \texttt{CAPS} section.
As each transcript evaluation takes approximately two hours of expert time,
we also explore automated evaluation using LLMs to complement our approach.
Prior research has shown that LLM-based evaluations produce results comparable to those of human experts,
supporting their effectiveness as a scalable evaluation method.\citep{Chiang2023}

\begin{table*}[htbp!]
    \centering 
    \caption{Detailed evaluation metrics and their principles.
        \textsc{Comp}: Comprehensiveness, \textsc{Appr}: Appropriateness,
        \textsc{CommS}: Communication Style, \textsc{Compl}: Completeness,
        \textsc{Faith}: Faithfulness.}
    \label{tab:evaluation_metrics}
    \begin{tblr}{
        colspec = {MMX[l,m]},
        cell{1}{3}={c},
        cell{2}{1}={r=3}{c},
        cell{5}{1}={r=4}{c},
        hline{1, Z} = {0.5pt, solid},
        hline{2, 5} = {0.3pt, solid},
        hline{3,4,6,7,8}={3}{dashed}
            }
                                  & \bf Metric & \bf Principles                                                                  \\
        \rtb{\texttt{AGENT}}      & \sc Comp   & {- Explores the symptom in depth                                                \\ - Asks quality and relevant follow-up questions }                                                                                                                          \\
                                  & \sc Appr   & {- Maintains focus on the symptom and diagnostically significant information    \\ - Recognize and response to emotional cues, such as acknowledging patient's distress or avoidance                          \\ - Creates a safe and supportive environment, such as showing understanding and empathy}                                    \\
                                  & \sc CommS  & {- Use clear and explanatory language to formulate the interview questions      \\ - Maintains smooth interview progresses and transitions between interview questions                                        \\ - Appropriately use clarifying questions, summaries and restatements to ensure accurate understanding of patient responses \\ - Adapts the dialogue to patient responses and emotional states }                                                                                                                        \\
        \rtb{\texttt{SIMULATION}} & \sc Compl  & {- Are simulated patient responses aligned with the original patient responses? \\ - Are simulated patient responses contextually appropriate to interview questions?\\ - Is the conversation flow between the clinician and simulated patient natural?
        }                                                                                                                        \\
                                  & \sc Appr   & {- Is the key information from the original transcript preserved accurately?    \\ - Are the simulated patient responses consistent with the original patient responses?\\ - Does the simulation avoid introducing hallucination or false information?
        }                                                                                                                        \\
                                  & \sc Faith  & {- Is the key information from the original transcript preserved accurately?    \\ - Are the simulated patient responses consistent with the original patient responses?\\ - Does the simulation avoid introducing hallucination or false information?
        }                                                                                                                        \\
                                  & \sc CommS  & {- Does the simulated patient response mirror the pace and language patterns?   \\ - Does the simulated patient response replicate the emotional states (e.g., distress or hesitation)?} \\
    \end{tblr}
\end{table*}

\subsubsection{Evaluation Metrics}
\label{sssec:metrics}
To ensure comprehensive assessment of our dialogue system,
we develop distinct evaluation metrics for both the agent generation and patient simulation components.
We employ a 5-point Likert scale ranging from -2 to 2 across all metrics,
with each metric comprising 2 to 4 specific evaluation principles.
Table~\ref{tab:evaluation_metrics} provides detailed descriptions of these principles.

For agent evaluation, we implement pairwise comparison between generated and original transcripts.
Scores from -2 to 2 indicate whether transcript A performs somewhat/significantly worse or better than transcript B.
Our dialogue evaluation framework encompasses three key metrics:
\textsc{Comprehensiveness}, which measures thoroughness of symptom exploration;
\textsc{Appropria\-teness}, which evaluates clinical effectiveness;
and \textsc{Communication Style}, which assesses interaction quality and dialogue flow from the agent perspective.
To standardize interpretation of comparative performance,
we establish performance thresholds: scores ranging from -0.3 to 0.3 indicate \textit{Equivalent Performance}
to the original transcripts, scores above 0.3 represent \textit{Enhanced Performance},
and scores below -0.3 signify \textit{Inadequate Performance}.


Patient simulation evaluation follows a different approach,
measuring evaluator agreement with specific questions under each metric.
Within this framework, scores of ±2 indicate strong agreement/disagreement,
while 0 represents a neutral assessment.
Similarly, we define three performance categories as \textit{Needs Improvement} (below 0.0),
\textit{Acceptable} (0.0 to +1.0), and \textit{Strong} Performance (above +1.0).


\begin{table}[!htbp]
    \centering 
    \caption{Evaluation results for 5 \texttt{CAPS} interview sections.
        \textbf{CE}: Conversation Expert, \textbf{DE}: Domain Expert,
        \textbf{LLM}: \texttt{claude-3-5-sonnet-20241022} and \texttt{gpt-4o}.
        \textsc{Avg}: average scores for all metrics.
        Scores are color-coded to reflect performance levels.
        For \texttt{Agent}: \cbox[colback=yellow9]{Equivalent}, \cbox[colback=green9]{Enhanced},
        \cbox[colback=red9]{Inadequate}.
        For \texttt{Simulation}: \cbox[colback=yellow9]{Acceptable}, \cbox[colback=green9]{Strong},
        \cbox[colback=red9]{Needs Improvement}. }
    \label{tab:evaluation_results}
    \begin{tblr}{
        colspec = {MMMMMM},
        cell{1}{3} = {c=2}{m},
        cell{1}{5} = {c=2}{m},
        cell{3}{1} = {r=4}{c},
        cell{7}{1} = {r=5}{c},
        cell{3-6}{3-4} = {yellow9},
        cell{7-8}{3-4} = {yellow9},
        cell{10-11}{3-4} = {yellow9},
        cell{8}{5} = {yellow9},
        cell{9}{6} = {yellow9},
        cell{3-6}{5-6} = {green9},
        cell{7-8}{6} = {green9},
        cell{10-11}{6} = {green9},
        cell{9}{3-4} = {red9},
        cell{7}{5} = {red9},
        cell{9-11}{5} = {red9},
        hline{1, Z} = {0.5pt, solid},
        hline{2} = {3-4}{0.1pt, solid, leftpos=-1, rightpos=-1, endpos},
        hline{2} = {5-6}{0.1pt, solid, leftpos=-1, rightpos=-1, endpos},
        hline{3, 7} = {0.3pt, solid}
            }
                                  &           &\bf Human&        & \bf LLM    &         \\
                                  &           & \bf CE & \bf DE & \bf Claude & \bf GPT \\
        \rtb{\texttt{AGENT}}      & \sc Comp  & -0.08  & -0.17  & 1.52       & 1.52    \\
                                  & \sc Appr  & -0.02  & -0.04  & 1.80       & 1.80    \\
                                  & \sc CommS & 0.20   & -0.20  & 1.72       & 1.96    \\
                                  & \sc Avg   & 0.03   & -0.14  & 1.68       & 1.76    \\
        \rtb{\texttt{SIMULATION}} & \sc Compl & 0.55   & 0.41   & -0.48      & 1.20    \\
                                  & \sc Appr  & 0.61   & 0.52   & 0.24       & 1.48    \\
                                  & \sc Faith & -0.39  & -0.33  & -0.20      & 0.48    \\
                                  & \sc CommS & 0.52   & 0.65   & -0.32      & 0.92    \\
                                  & \sc Avg   & 0.32   & 0.31   & -0.19      & 1.02    \\
    \end{tblr}
\end{table}

\subsubsection{Quantitative Results}
\label{sssec:eval_results}


To achieve the optimal level of granularity in evaluating dialogue quality,
we implement a symptom-based evaluation for the \texttt{CAPS} section.
Utterance-level evaluation is overly fine-grained and overlooks broader dialogue patterns,
while transcript-level evaluation is too coarse to capture topic-specific performance.
Our symptom-based approach aligns with the structural orgnization of the \texttt{CAPS} section,
which systematically assesses various PTSD-related symptoms.
Specifically, we segment each transcript by variables and group these variables into 25 symptom clusters,
each corresponding to a distinct PTSD symptom.
Each cluster contains 3 to 4 related variables,
providing sufficient context for meaningful evaluation while maintaining focus on symptom-specific criteria.
Table~\ref{tab:evaluation_results} presents the evaluation results for the 5 sampled transcripts.
For automatic evaluation, we additionally incorporate GPT alongside Claude to mitigate potential self-evaluation bias,
as LLMs may favor their own generations.\citep{Panickssery2024}
Both models are prompted as evaluators using the same instructions,
with the temperature parameter set to 0.

For agent evaluation, human evaluators
rate the system performance at 0.03 and -0.14 on average,
indicating equivalent performance to the original transcripts in general.
Scores for \textsc{Comprehensiveness} and \textsc{Appropriateness} show consistency between both experts.
The \textsc{Communication Style} metric receives a higher score from the conversation expert at 0.20, while the domain expert assesses the same metric at -0.20.

In contrast, the LLM evaluators consistently rated agent performance within the \textit{Enhanced Performance} range,
with average scores of 1.68 and 1.76.
This notable discrepancy between human and LLM evaluations underscores the limitations of LLMs
in accurately assessing dialogue quality in clinical interviews.
These findings suggest that LLMs, especially when used in isolation,
are currently unreliable for large-scale evaluation in this domain.

For patient simulation, both human evaluators provide positive assessments,
with average scores of 0.32 and 0.31, falling in the \textit{Acceptable} range.
Most metrics except for \textsc{Faithfulness} receive scores that exceed the acceptable threshold.
Both human evaluators give negative scores for \textsc{Faithfulness} at -0.39 and -0.33,
suggesting that the LLM struggles to preserve all key information presented in the original transcripts without hallucination.

The LLM evaluators demonstrate significant divergence in their assessment of patient simulation (1.02 from GPT versus -0.19 from Claude).
This inconsistency further highlights the unreliability of LLMs as sole evaluators and
reinforces the necessity of human evaluation in this context.

\begin{table}[htbp!]
  \centering 
  \caption{Statistics of the sampled original transcripts (\texttt{Diaglogue 1-5}).
    \textbf{Utter}: utterance counts, \textbf{Tok}: token counts,
    \textbf{Ques}: the number of questions in each dialogue that match the predefined \texttt{CAPS} interview questions,
    \textbf{Score}: the average of semantic similarity and fuzz matching scores for these matched questions.}
  \label{tab:dialogue_stats}
  \begin{tblr}{
    colspec = {MMMMMMM},
    hline{1, Z} = {0.5pt, solid},
        hline{3}={0.3pt,solid},
        hline{2}={2-3}{0.1pt,leftpos=-1,rightpos=-1,endpos},
        hline{2}={4-5}{0.1pt,leftpos=-1,rightpos=-1,endpos},
        hline{2}={6-7}{0.1pt,leftpos=-1,rightpos=-1,endpos},
        cell{1}{2}={c=2}{c},
        cell{1}{4}={c=2}{c},
        cell{1}{6}={c=2}{c},
      }
    & \bf Clinician &       & \bf Patient &       & \bf Match &       \\
    & Utter         & Tok   & Utter       & Tok   & Ques      & Score \\
    \tt D1 & 162           & 3,658 & 142         & 1,605 & 83        & 1.52  \\
    \tt D2 & 159           & 3,558 & 133         & 3,434 & 63        & 1.61  \\
    \tt D3 & 147           & 3,555 & 127         & 3,457 & 73        & 1.49  \\
    \tt D4 & 126           & 2,837 & 114         & 4,567 & 67        & 1.64  \\
    \tt D5 & 82            & 2,575 & 65          & 1,401 & 46        & 1.59  \\
  \end{tblr}
\end{table}

\begin{figure}[htbp!]
  \centering
  \subfloat[Agent generation evaluation results.]{
    \includegraphics[width=\columnwidth]{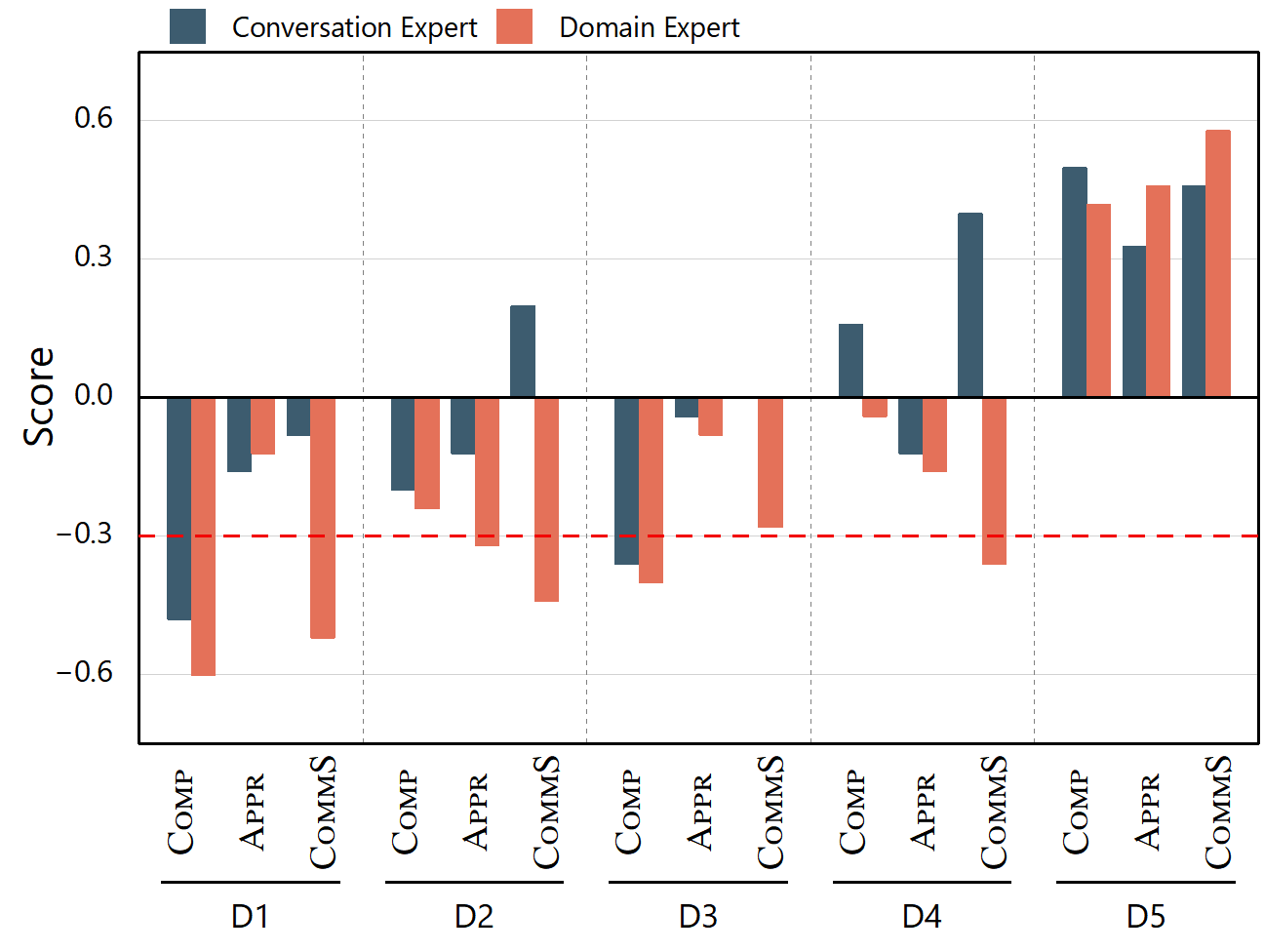}
    \label{subfig:dialogue_detailed}
  } \\
  \subfloat[Patient simulation evaluation results.]{
    \includegraphics[width=\columnwidth]{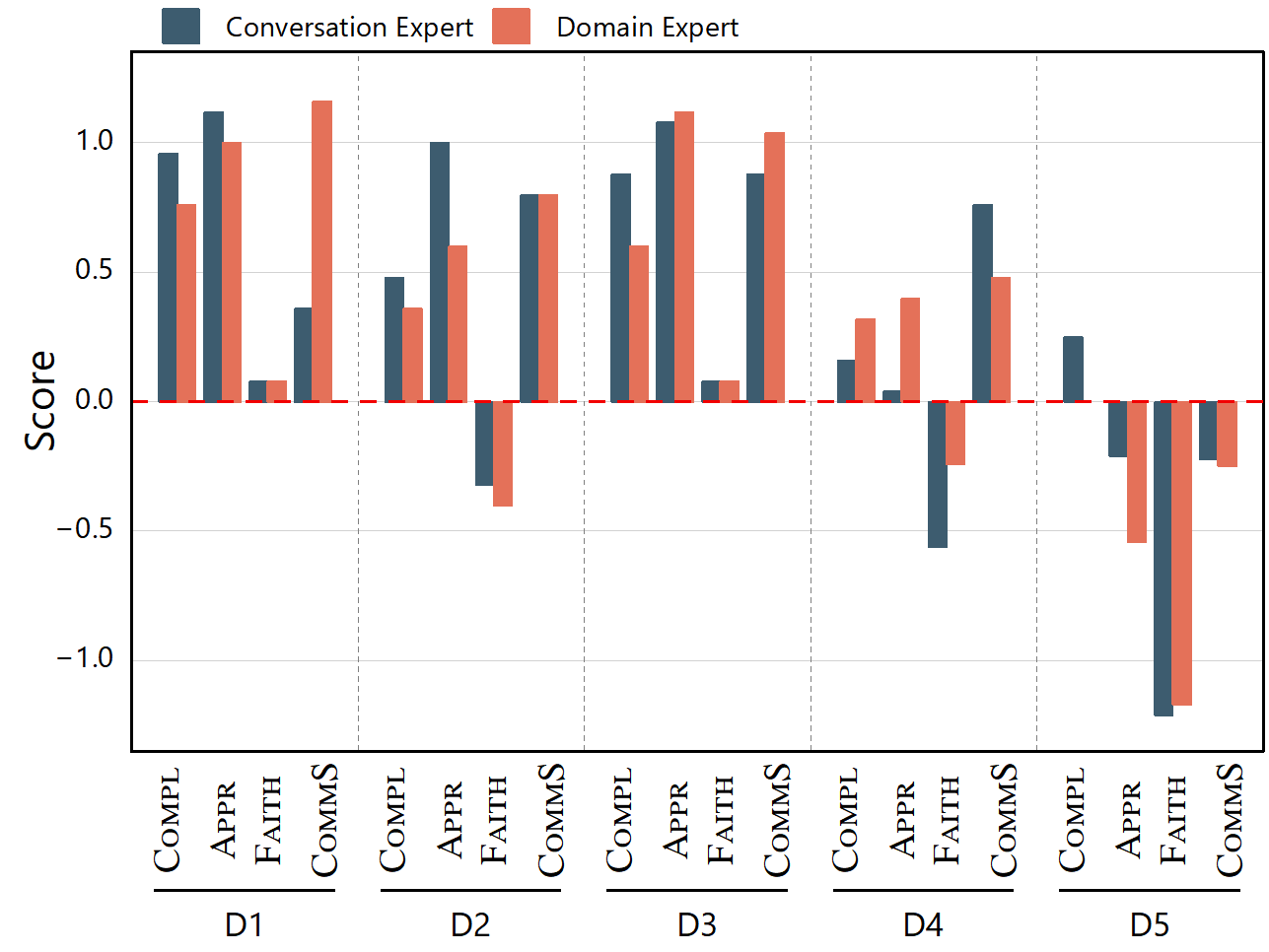}
    \label{subfig:simulation_detailed}}
  \caption{Evalutaion results across 5 dialogues (\texttt{Dialogue 1-5}).
    The x-axis displays evaluation metrics, grouped by dialogue ID.
    The y-aixs shows evaluation scores, with a red dashed reference line
    marking the lower threshold of \textit{Equivalent Performance} for the agent
    and \textit{Acceptable} for simulation.
    Note that the reference line for Patient Simulation overlaps the 0.0 value line.}
  \label{fig:detailed_eval_by_dialogue}
\end{figure}

\section{Discussion}
\label{sec:analysis}
We conduct a comprehensive analysis of the generated and original transcripts through side-by-side comparison.
Table~\ref{tab:evaluation_results} demonstrates general alignment in evaluations
from both the conversation expert and domain expert,
as evidenced by their similar scoring ranges for dialogue and simulation performance.
We further examine the statistical characteristics of the original transcripts (Table~\ref{tab:dialogue_stats})
alongside the evaluation scores for each dialogue (Figure~\ref{fig:detailed_eval_by_dialogue}).
Several meaningful patterns are identified to provide greater insight into the system performance beyond the aggregate metrics.

\subsubsection{Agent Communication Style}
The evaluation discrepancy in agent \textsc{Communication Style}
(-0.2 from the domain expert versus 0.2 from the conversation expert)
stems from their different professional perspectives.
Two common sources of this divergence are:
\begin{enumerate}
  \setlength\itemsep{-0.1em}
  \item \texttt{ACK} (Acknowledgement), \texttt{EMP} (Empathy), and \texttt{VAL} (Validation)
        before transitioning to the next criterion.
        The conversation expert views these elements as promoting smoother transitions 
        and demonstrating the system's ability to integrate prior context. 
        In contrast, the domain expert notes that such responses can distract from the diagnostic focus of the new criterion item.
  \item Agent interpretation misalignment. 
   The agent occasionally extrapolates beyond what the patient actually stated, 
   which compromises diagnostic accuracy from a clinical perspective. 
   The domain expert observed exaggerated interpretations in Dialogues 1–4, consistent with lower evaluation scores.
        Figure~\ref{subfig:agent_CommS} provides a clear example of both scenarios.
\end{enumerate}

\subsubsection{Agent Appropriateness} 
\begin{figure}[htbp!]
  \centering
  \subfloat[Inappropriate communication style.]{
    \includegraphics[width=\columnwidth]{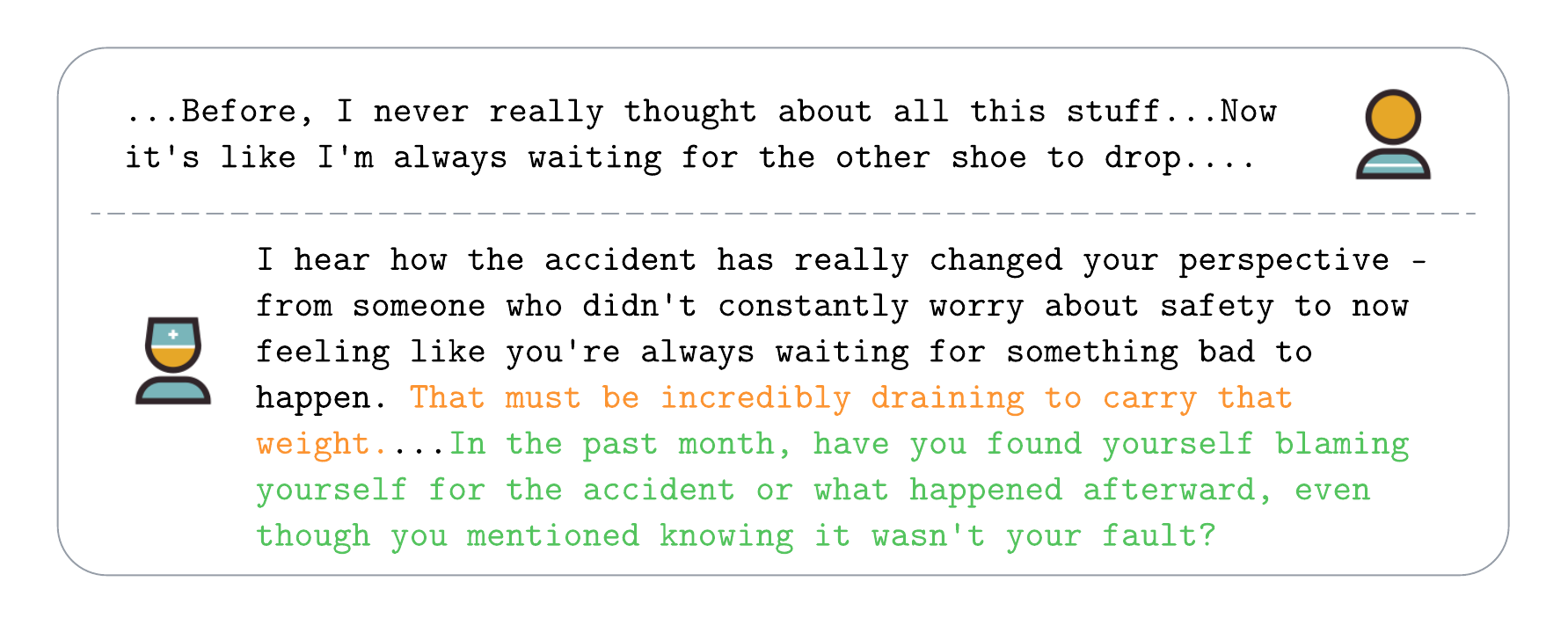}
    \label{subfig:agent_CommS}} \\
  \subfloat[Inappropriate interview question.]{
    \includegraphics[width=\columnwidth]{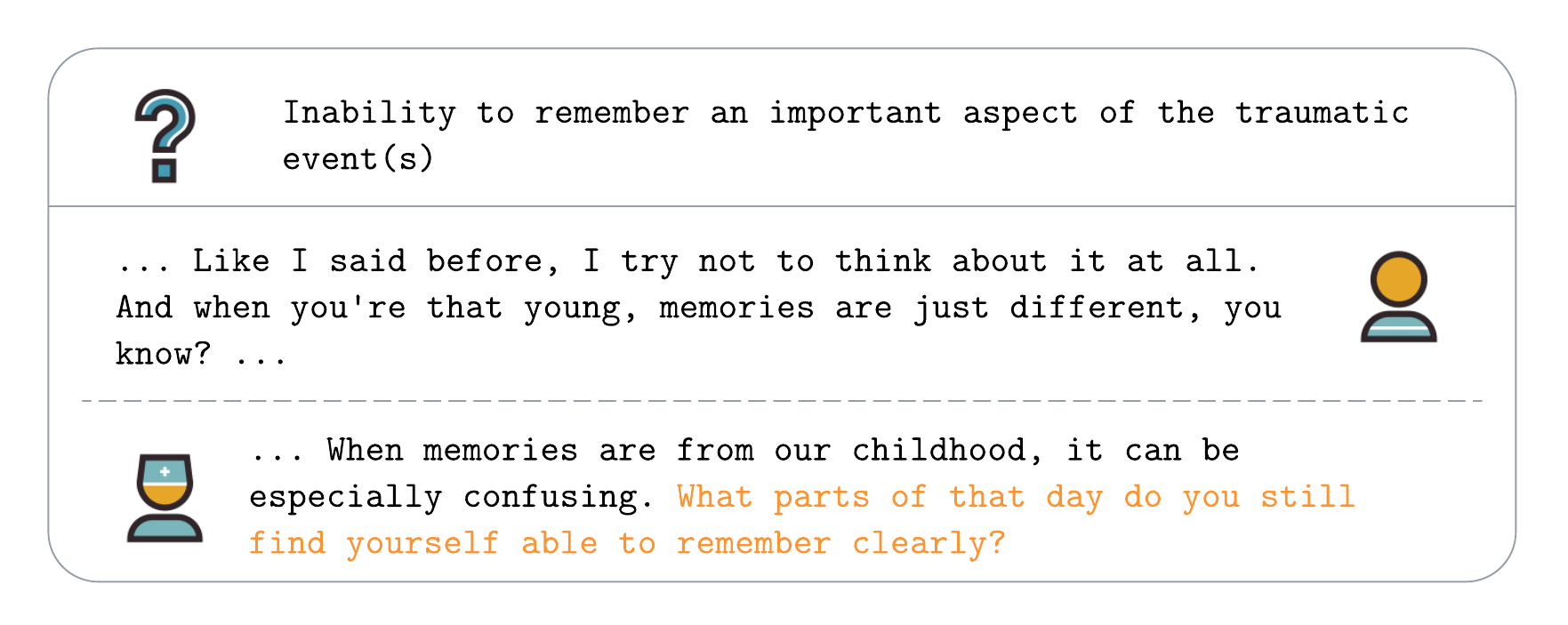}
    \label{subfig:appropriateness_ex}
  }
  \caption{Examples of inappropriate \textsc{Communication Style} and interview question.
    Exaggerated interpretation and unnecessary question are highlighted in orange.
    Next \texttt{IS} question is highlighted in green. Question icon indicates the PTSD symptom to assess.}
  \label{fig:analysis_example}
\end{figure}

In addition to predefined \texttt{CAPS} interview questions,
clinicians employ tailored open-end clarifying questions (\texttt{CQ}) to elicit diagnostic details
based on individual patient experiences.
Our analysis reveals that the system occasionally requests information that,
while conversationally relevant to traumatic events, extends beyond diagnostic necessity.
As shown in Figure~\ref{subfig:appropriateness_ex}, the highlighted question is inappropriate
because it prompts the patient to recall unnecessary details,
whereas the diagnostic focus is on memory impairment
rather than the specifics of what can be remembered.
Although the question may seem relevant from a conversational standpoint,
it deviates from the clinical objective.


\subsubsection{Conversational Nuance} Our system performs well in generating correct DA tags for acknowledgement and validation.
However, two key challenges to the LLM are identified during the analysis:

\begin{enumerate}
  \setlength\itemsep{-0.1em}
  \item Lengthy \texttt{ACK} and \texttt{VAL}. The LLM tends to generate excessively long content for \texttt{ACK} and
        \texttt{VAL}, even when explicitly instructed to be concise.
  \item Difficulty replicating real-life conversational cues. In real-life interactions,
        clinicians often acknowledge the previous response with brief phrases like \textit{yeah} or \textit{alright},
        potentially receiving minimal reciprocal responses from patients,
        followed by a short pause before proceeding to the next question.
        While humans intuitively understand this exchange as a cue to move forward,
        the LLM struggles to replicate this nuanced conversational rhythm.
\end{enumerate}

\subsubsection{Simulation Faithfulness} Hallucination remains a known challenge for LLMs.
The negative evaluation scores of the patient simulation can be attributed to:

\begin{enumerate}
  \setlength\itemsep{-0.1em}
  \item Missing information in transcripts.
        If the relevant information to answer the interview question is missing
        or insufficient in the original transcripts,
        the LLM tends to assume an answer and fabricate supporting details to justify that assumption.
  \item Excessive inference. The model may deviate from the original transcription
        due to hallucination or overgeneralization.
        For example, original phrases such as
        \textit{going out to eat, hanging out with family} and
        \textit{using your hands to create things} are inaccurately expanded to
        \textit{going hiking, meeting friends for coffee, working in my garden}.
\end{enumerate}

\subsubsection{Transcript Quality} Since our evaluation method relies on pairwise comparison,
the quality of original transcripts influences the evaluation scores.
Figure~\ref{fig:detailed_eval_by_dialogue} reveals a general negative correlation between agent and simulation scores,
where higher agent scores corresponds to lower simulation scores.
For instance, in \texttt{D1}, the transcript contains the highest clinician utterance count
and question matching count among the 5 Diaglogues.
This suggests that the clinician in \texttt{D1} asks more optional interview questions in \texttt{CAPS} section
to gather detailed information about the patient's symptoms.
As a result, \texttt{D1} is the most comprehensive transcript in terms of clinician responses,
which explains its low agent \textsc{Comprehensiveness} score.
The more exhaustive the orginal dialouge, the harder it is for the system to match its depth.
Conversely, \texttt{D1} achieves positive simulation \textsc{Faithfulness} socre
since most system-generated questions could be mathced with relevant information in the orignal transcipt,
given its extensive coverage of optional interview questions.


\section{Conclusion}
\label{sec:conclusion}

In this paper, we introduce TRUST, a dialogue system designed to conduct formal diagnostic interviews for PTSD assessment.
To enhance control over the decision-making process during response generation, we develop a DA schema for clinician responses.
As this schema does not contain condition-specific tags, it can be adapted to other clinical interview scenarios.
To evaluate the system's effectiveness, we integrate patient simulation as user input,
leveraging original transcripts for robust testing.
Expert evaluations from both conversation and domain specialists demonstrate that
TRUST achieves performance comparable to clinicians on average.
This work bridges a critical gap in mental health dialogue systems and lays the foundation for
expanding structured diagnostic interviews to broader mental health conditions.


\section*{Ethics Statements}
The data used in this paper was collected with informed consent approved
by the Institutional Review Board (IRB) and Research Oversight Committee.
The authors and clinicians involved in the research have passed
Research, Ethics, Compliance, and Safety Training through Collaborative Institutional Training Initiative (CITI Program).
For the use of LLMs, this study exclusively employs anonymized interviews,
ensuring the confidentiality and privacy of all participants.

\section*{Acknowledgments}
\label{sec:acknowledgments}

We gratefully acknowledge the support of the DooGood Foundation. 
Any opinions, findings, and conclusions or recommendations expressed in this material 
are those of the authors and do not necessarily reflect the views of the DooGood Foundation.

We would like to thank Vasiliki Michopoulos, Jennifer Stevens, Rebecca Hinrichs, Negar Fani, 
Angelo Brown, Natalie Merrill, Rebecca Lipschutz, and the entire Grady Trauma Project team 
for their assistance and participants for their study engagement and time.

We also thank Tung Dinh for his contributions to drafting the DA annotation schema 
and providing manual annotations for the DA tags.

\bibliographystyle{unsrtnat}
\bibliography{reference}




\end{document}